%% file: sample-aamas19.tex
\documentclass[sigconf]{aamas}  

\usepackage{booktabs}

\setcopyright{ifaamas}  
\acmDOI{doi}  
\acmISBN{}  
\acmConference[AAMAS'19]{Proc.\@ of the 18th International Conference on Autonomous Agents and Multiagent Systems (AAMAS 2019), N.~Agmon, M.~E.~Taylor, E.~Elkind, M.~Veloso (eds.)}{May 2019}{Montreal, Canada}  
\acmYear{2019}  
\copyrightyear{2019}  
\acmPrice{}  

\DeclareMathOperator*{\argmax}{arg\,max}

\renewcommand\footnotetextcopyrightpermission[1]{} 
\begin{document}

\title{Learning Self-Game-Play Agents for Combinatorial Optimization Problems}  




%
\author{Ruiyang Xu}
\affiliation{%
 \institution{Khoury College of Computer Sciences \\ Northeastern University}
 \city{Boston} 
 \state{MA} 
 \postcode{02115}
}
\email{ruiyang@ccs.neu.edu}

\author{Karl Lieberherr}
\affiliation{%
 \institution{Khoury College of Computer Sciences\\ Northeastern University}
 \city{Boston} 
 \state{MA} 
 \postcode{02115}
}
\email{lieber@ccs.neu.edu}
%
%
%
%
%
%

\newcommand\ZG{Zermelo Gamification}
\begin{abstract}  
Recent progress in reinforcement learning (RL) using self-game-play has shown remarkable performance on several board games (e.g., Chess and Go) as well as video games (e.g., Atari games and Dota2). It is plausible to consider that RL, starting from zero knowledge, might be able to gradually approximate a winning strategy after a certain amount of training. In this paper, we explore neural Monte-Carlo-Tree-Search (neural MCTS), an RL algorithm which has been applied successfully by DeepMind to play Go and Chess at a super-human level. We try to leverage the computational power of neural MCTS to solve a class of combinatorial optimization problems. Following the idea of Hintikka's Game-Theoretical Semantics, we propose the \ZG\ (ZG) to transform specific combinatorial optimization problems into Zermelo games whose winning strategies correspond to the solutions of the original optimization problem. A specially designed neural MCTS algorithm is also provided to train Zermelo game play agents. We use a prototype problem for which the ground-truth policy is efficiently computable to demonstrate that ZG is promising.
\end{abstract}

%

\keywords{Reinforcement Learning; neural MCTS; Self-game-play; Combinatorial Optimization; Tabula rasa}  

\maketitle


\input{samplebody-conf}


\bibliographystyle{ACM-Reference-Format}  
\bibliography{sample-bibliography}  

\end{document}

%% file: samplebody-conf.tex

\section{Introduction}\label{introduction}
The past several years have witnessed the progress and success of reinforcement learning (RL) in the field of game-play. The combination of classical RL algorithms with newly developed deep learning techniques gives stunning performance on both traditional simple Atari video games \cite{dqn} and modern complex RTS games (like Dota2 \cite{ppo}), and even certain hard board games like Go and Chess \cite{alpha0}. One common but outstanding feature of those learning algorithms is the tabula-rasa style of learning. In terms of RL, all those algorithms are model-free\footnote{Considering the given problem as an MDP (Markov Decision Process), the learning algorithm doesn't have to know in advance the transition probabilities and rewards after each action is taken} and learn to play the game with zero knowledge (except the game rules) in the beginning. Such tabula-rasa learning can be regarded as an approach towards a general artificial intelligence.

Although there are lots of achievements in game, there is little literature on how to apply those techniques to general problems in other domains. It is tempting to see whether those game-play agents' superhuman capability can be used to solve problems in other realms. In this work, we transform a family of combinatorial optimization problems into games via a process called \ZG, so that an AlphaZero style (i.e., neural MCTS \cite{alpha0,silver2018general}) game-play agent can be leveraged to play the transformed game and solve the original problem. Our experiment shows that the two competitive agents gradually, but with setbacks, improve and jointly arrive at the optimal strategy. The tabula-rasa learning converges and solves a non-trivial problem, although the Zermelo game is fundamentally different from Go and Chess. The trained game-play agent can be used to approximate\footnote{For problems with small sizes, one can achieve an optimal solution by providing the learning algorithm enough computing resources.} the solution (or show the non-existence of a solution) of the original problem through competitions against itself based on the learned strategy.

We make three main contributions: 1. We introduce the \ZG\, a way to transform combinatorial problems to Zermelo games using a variant of Hintikka's Game-Theoretical Semantics \cite{semgame}; 2. We implemented a modification of the neural MCTS algorithm\footnote{Our implementation is based on an open-source, lightweight framework: AlphaZeroGeneral, https://github.com/suragnair/alpha-zero-general} designed explicitly for those Zermelo games; 3. We experiment with a prototype problem (i.e., $HSR$). Our result shows that, for problems under a certain size, the trained agent does find the optimal strategy, hence solving the original optimization problem in a tabula-rasa style.

The remainder of this paper is organized as follows.
Section \ref{preliminaries} presents essential preliminaries on neural MCTS and combinatorial optimization problems which we are interested in. Section \ref{game} introduces Zermelo game and a general way to transform the given type of combinatorial optimization problems into 
Zermelo games, where we specifically discuss our prototype problem $HSR$.
Section \ref{experiment} gives our correctness measurement and presents experimental results. \ref{discuss} and \ref{conclusion} made a discussion and conclusions.
%

\section{Preliminaries}\label{preliminaries}

\subsection{Monte Carlo Tree Search}\label{mcts}
The PUCT (Predictor +
UCT) algorithm implemented in AlphaZero \cite{alpha0,gochessshogi} is essentially a neural MCTS algorithm which uses PUCB Predictor +
UCB \cite{Rosin2011} as its confidence upper bound \cite{uct,ucb1} and uses the neural prediction $P_{\phi}(a|s)$ as the predictor. The algorithm usually proceeds through 4 phases during each iteration:
\begin{enumerate}
\item{SELECT: }
At the beginning of each iteration, the algorithm selects a path from the root (current game state) to a leaf (either a terminal state or an unvisited state) in the tree according to the PUCB (see \cite{alpha_go} for detailed explanation for terms used in the formula). Specifically, suppose the root is $s_0$, we have \footnote{Theoretically, the exploratory term should be $\sqrt{\frac{\sum_{a'}N(s_{i-1},a')}{N(s_{i-1},a)+1}}$, however, the AlphaZero used the variant $\frac{\sqrt{\sum_{a'} N(s_{i-1},a')}}{N(s_{i-1},a)+1}$ without any explanation. We tried both in our implementation, and it turns out that the AlphaZero one performs much better.}:
$$a_{i-1}=\argmax_a\left[Q(s_{i-1},a)+cP_\phi(a|s_{i-1})\frac{\sqrt{\sum_{a'} N(s_{i-1},a')}}{N(s_{i-1},a)+1}\right]$$
$$Q(s_{i-1},a) = \frac{W(s_{i-1},a)}{N(s_{i-1},a)+1}$$
$$s_i=next(s_{i-1},a_{i-1})$$
\item{EXPAND: }
Once the select phase ends at a non-terminal leaf, the leaf will be fully expanded and marked as an internal node of the current tree. All its children nodes will be considered as leaf nodes during next iteration of selection.
\item{ROLL-OUT: }
Normally, starting from the expanded leaf node chosen from previous phases, the MCTS algorithm uses a random policy to roll out the rest of the game \cite{mcts_survey}. The algorithm simulates the actions of each player randomly until it arrives at a terminal state which means the game has ended. The result of the game (winning information or ending score) is then used by the algorithm as a result evaluation for the expanded leaf node.

However, a random roll-out usually becomes time-consuming when the tree is deep. A neural MCTS algorithm, instead, uses a neural network $V_{\phi}$ to predict the result evaluation so that the algorithm saves the time on rolling out.   

\item{BACKUP: }
This is the last phase of an iteration where the algorithm recursively backs-up the result evaluation in the tree edges. Specifically, suppose the path found in the Select phase is $\{(s_0,a_0),(s_1,a_1),...(s_{l-1},a_{l-1}),(s_l,\_)\}$. then for each edge $(s_i,a_i)$ in the path, we update the statistics as:
$$W^{new}(s_i,a_i)=W^{old}(s_i,a_i)+V_{\phi}(s_l)$$
$$N^{new}(s_i,a_i)=N^{old}(s_i,a_i)+1$$
However, in practice, considering the +1 smoothing in the expression of Q, the following updates are actually applied:
$$Q^{new}(s_i,a_i)=\frac{Q^{old}(s_i,a_i)\times N^{old}(s_i,a_i)+V_{\phi}(s_l)}{N^{old}(s_i,a_i)+1}$$
$$N^{new}(s_i,a_i)=N^{old}(s_i,a_i)+1$$
Once the given number of iterations has been reached, the algorithm returns a vector of action probabilities of the current state (root $s_0$). And each action probability is computed as $\pi(a|s_0)=\frac{N(s_0,a)}{\sum_{a'}N(s_0,a')}$. The real action played by the MCTS is then sampled from the action probability vector $\pi$. In this way, MCTS simulates the action for each player alternately until the game ends, this process is called MCTS simulation (self-play).
\end{enumerate}

\subsection{Combinatorial Optimization Problems}\label{cop}
The combinatorial optimization problems studied in this paper can be described with the following logic statement:
$$\exists n:\{G(n)\wedge(\forall n'>n~\neg G(n'))\}$$
$$G(n):=\forall x~\exists y:\{F(x,y;n)\}$$
or
$$G(n):=\exists y~\forall x:\{F(x,y;n)\}$$
In this statement, $n$ is a natural number and $x,y$ can be any instances depending on the concrete problem. $F$ is a predicate on $n,x,y$. Hence the logic statement above essentially means that there is a maximum number $n$ such that for all $x$, some $y$ can be found so that the predicate $F(x,y;n)$ is true. 
Formulating those problems as interpreted logic statements is crucial to transforming them into games (Hintikka \cite{semgame}). In the next section, we will introduce our gamification method in details. 

\section{\ZG}\label{game}
\subsection{General Formulation}\label{general_form}
%
%
We introduce the \ZG\ (ZG) to transform a combinatorial optimization problem into a Zermelo game that is fit for being used by a specially designed neural MCTS algorithm to find a winning strategy. The winning strategy can be used to find a solution to the original combinatorial problem. We will illustrate the \ZG\ by deriving a prototype game: $HSR$.

\begin{figure}
\centering
\includegraphics[width=0.4\textwidth]{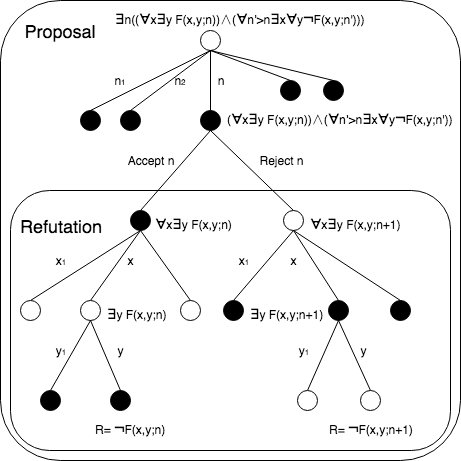}
\caption{ An overall Zermelo game where white nodes stand for the P's turn and black nodes stand for the OP's turn. A role-flip happened after OP's rejecting of $n$. The refutation game can be treated uniformly where, depending on the order of the quantifiers, the OP takes the first move. The OP wins if and only if the P fails to find any $y$ to make $F(x,y;n)$ holds, hence the game result $R=\neg F(x,y;n)$.}
\label{Zermelo_game}
\end{figure}

A Zermelo game is defined to be a two-player, finite, and perfect information game with only one winner and loser, and during the game, players move alternately (i.e., no simultaneous move). Leveraging the logic statement (see section \ref{cop}) of the problem, the 
Zermelo game is built on the Game-Theoretical Semantic approach (Hintikka \cite{semgame}). 
We introduce two roles: the Proponent (P), who claims that the statement is true, and the Opponent (OP), who argues that the statement is false. The original problem can be solved if and only if the P can propose some optimal number $n$ so that a perfect OP cannot refute it.
To understand the game mechanism, let's recall the logic statement in section \ref{cop}, which implies the following
Zermelo game (Fig. \ref{Zermelo_game}):
\begin{enumerate}
\item Proposal Phase:
in the initial phase of the Zermelo game player P will propose a number $n$. Then the player OP will decide whether to accept this $n$, or reject it. OP will make his decision based on the logic statement:
$A\wedge B, A:=G(n), B:=\forall n'>n~\neg G(n')$. 
Specifically, the OP tries to refute the P by attacking either on the statement $A$ or $B$. The OP will accept $n$ proposed by the P if she confirms $A=False$. The OP will reject $n$ if she is unable to confirm $A=False$. In this case, The OP treats $n$ as non-optimal, and proposes a new $n'>n$ (in practice, for integer $n$, we take $n'=n+1$) which makes $B=False$. To put it in another way, $B=False$ implies $\neg B=True$ which also means that the OP claims $G(n')$ holds. Therefore, the rejection can be regarded as a role-flip between the two players. To make the Zermelo non-trivial, in the following game, we require that the P has to accept the new $n'$ and tries to figure out the corresponding $y$ to defeat the OP. Notice that since this is an adversarial game, the OP will never agree with the P (namely, the OP will either decide that the $n$ is too small or too large because the OP has to regard the move made by the P as incorrect). Therefore, the OP is in a dilemma when the P is perfect, i.e., the P chooses the optimal $n$.

\item Refutation Phase: 
it is the phase where the two players search for evidence and construct strategies to attack each other or defend themselves. Generally speaking, regardless of the role-flip, we can treat the refutation game uniformly: the P claims $G(n)$ holds for some $n$, the OP will refute this claim by giving some instances of $x$ (for existential quantifier) so that $\neg G(n)$ holds. If the P successfully figures out the exceptional $y$ (for universal quantifier) which makes $F(x,y;n)$ hold, the OP loses the game; otherwise, the P loses. 
\end{enumerate}

The player who takes the first move is decided by order of the quantifiers, namely, for $G(n):=\exists x\forall y:\{F(x,y;n)\}$ the P will take the first move; for $G(n):=\forall y\exists x:\{F(x,y;n)\}$ the OP will take the first move. The game result is evaluated by the truth value of $F(x,y;n)$, specifically, if the P takes the last move then she wins when $F(x,y;n)$ holds; otherwise, if the OP makes the last move then she wins when $F(x,y;n)$ doesn't hold. It should be noticed that the OP is in a dilemma when the P is perfect.


\subsection{HSR Problem}\label{hsr_game}

In this section, we first introduce our prototype, the Highest Safe Rung ($HSR$) problem. Then, we will see how to perform Zermelo gamification on it. 

The $HSR$ problem can be described as follows: consider throwing jars from a specific rung of a ladder. The jars could either break or not. If a jar is unbroken during a test, it can be used next time. A highest safe rung is a rung that for any test performed above it, the jar will break. Given $k$ identical jars and $q$ test chances to throw those jars, what is the largest number of rungs a ladder can have so that there is always a strategy to locate the highest safe rung with at most $k$ jars and $q$ tests? 

To formulate $HSR$ problem in predicate logic, we utilize the recursive property. Notice that, after performing a test, depends on whether the jar is broken or not, the highest safe rung should only be located either above the current testing level, or below or equal to the current testing level. This fact means that the next testing level should only be located in the upper partial ladder or the lower partial ladder. Therefore, the original problem can be divided into two sub-problems. We introduce the predicate $G_{k,q}(n)$ which means there is a strategy to find the highest safe rung on an $n$-level ladder with at most $k$ jars and $q$ tests. Specifically, using the recursive property we have mentioned, $G_{k,q}(n)$ can be written as:
$$G_{k,q}(n)=\exists\ 0<m\le n\ :\{G_{k-1,q-1}(m-1)\wedge G_{k,q-1}(n-m)\}$$
$$G_{k,q}(0)=True,G_{0,q}(n)=False,G_{k,0}(n)=False,G_{0,0}(n)=False$$
$$n>0,k>0,q>0$$
This formula can be interpreted as following: if there is a strategy to locate the highest safe rung on an $n$-level ladder, then it must tell you a testing level $m$ so that, no matter the jar breaks or not, the strategy can still lead you to find the highest safe rung in the following sub-problems. More specifically, for sub-problems, we have $G_{k-1,q-1}(m-1)$ if the jar breaks, that means we only have $k-1$ jars and $q-1$ tests left to locate the highest safe rung in the lower partial ladder (which has $m-1$ levels). Similarly, $G_{k,q-1}(n-m)$ for upper partial ladder. Therefore, the problem is solved recursively, and until there is no ladder left, which means the highest safe rung has been located, or there is no jars/tests left, which means one has failed to locate the highest safe rung. With the notation of $G_{k,q}(n)$, the $HSR$ problem now can be formulated as:
$$HSR_{k,q}=\exists n:\{G_{k,q}(n)\wedge(\forall n'>n~\neg G_{k,q}(n'))\}$$

Next, we show how to perform Zermelo gamification on the $HSR$ problem. Notice that the expression of $G_{k,q}(n)$ is slightly different with the ones used in section \ref{cop}: there is no universal quantifiers in the expression. To introduce the universal quantifier, we regard the environment as another player who plays against the tester so that, after each testing being performed, the environment will tell the tester whether the jar is broken or not. In this way, locating the highest safe can be formulated as following:
$$G_{k,q}(n)=\begin{cases} \mbox{True},\ \mbox{if } n=0 \\ \mbox{False},\ \mbox{if } n>0\wedge (k=0\vee q=0) 
\\\exists m\in [1...n]\ \forall a\in \text{Bool}:\\
\{(a\rightarrow G_{k-1,q-1}(m-1))\wedge (\neg a\rightarrow G_{k,q-1}(n-m))\} 
\end{cases}$$

\begin{figure}[ht]
\centering
\includegraphics[width=0.4\textwidth]{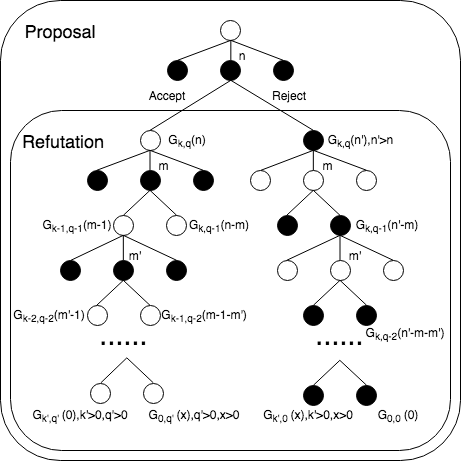}
\caption{The Zermelo gamification of $HSR$ problem. The game recursively played between two players until the highest safe rung being loacted or all resources have been used up.}
\label{hsr_game0_tree}
\end{figure}

Now, with the formula above, one can perform the standard Zermelo gamification (section \ref{general_form}) to get corresponding Zermelo game (Fig. \ref{hsr_game0_tree}). Briefly speaking, the tester now becomes the P player in the game, and the environment becomes OP. In the proposal phase, P will propose a number $n$ for which P thinks it is the largest number of levels a ladder can have so that she can locate any highest safe rung using at most $k$ jars and $q$ tests. The OP will decide whether to accept or reject this proposal by judging whether $n$ is too small or too large. In the refutation phase, P and OP will give a sequence of testing levels and testing results alternately, until the game ends. In this game, both P and OP will improve their strategy during the game so that they always play adversarial against each other and adjust one's strategy based on the reaction from the other one. 

\begin{figure}
\centering
\includegraphics[width=0.4\textwidth]{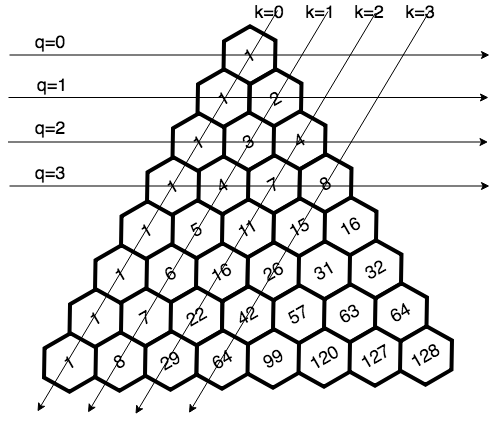}
\caption{Theoretical values for maximum $n$ in $HSR$ problem with given $k,q$, which can be represented as a Bernoulli's Triangle.}
\label{exttriangle}
\end{figure}

It should be mentioned that due to $HSR$ problem, as a prototype to test our idea, itself is not a hard problem, the solution for the $HSR$ problem can be computed and represented efficiently with a Bernoulli's Triangle (Fig. \ref{exttriangle}). We use the notation $N(k,q)$ to represent the solution for $HSR$ problem given $k$ jars and $q$ tests. In other words, $G_{k,q}(N(k,q))\wedge(\forall n'>N(k,q)~\neg G_{k,q}(n'))$ always holds.

\section{Experiment}\label{experiment}
\subsection{Neural MCTS implementation}
In this section, we will discuss our neural MCTS implementation on the $HSR$ game. Since the Zermelo game has two phases and the learning tasks are quite different between these two phases, we applied two independent neural networks to learn the proposal game and refutation game respectively. The neural MCTS will access the first neural network during the proposal game and then the second neural network during the refutation game. There are also two independent replay buffers which store the self-play information generated from each phase, respectively. 

Our neural network consists of four layers of 1-D convolution neural networks and two dense layers. The input is a tuple $(k,q,n,m,r)$ where $k,q$ are resources, $n$ is the number of rungs on the current ladder, $m$ is the testing point, and $r$ indicates the current player. The output of the neural network consists of two vectors of probabilities on the action space for each player as well as a scalar as the game result evaluation.

During each iteration of the learning process, there are three phases: 1. 100 episodes of self-play will be executed through a neural MCTS using the current neural network. Data generated during self-play will be stored and used for the next phase. 2. the neural networks will be trained with the data stored in the replay buffer. And 3. the newly trained neural network and the previous old neural network are put into a competition to play with each other. During the competition phase, the new neural network will first play as the OP for 20 rounds, and then it will play as the P for another 20 rounds. We collect the correctness data for both of the neural networks during each iteration. \footnote{It should be mentioned that the arena phase can be used only to obtain experimental data while the model can be continuously updated without the arena phase, as AlphaZero.}

We shall mention that since it is highly time-consuming to run a complete Zermelo game on our machines, to save time and as a proof of concept, we only run the entire game for $k=7,q=7$ and $n\in [1...130]$. Nevertheless, since the refutation game, once $n$ is given, can be treated independently from the proposal game, we run the experiment on refutation games for various parameters.

\subsection{Correctness Measurement}
Informally, an action is correct if it preserves a winning position. It is straightforward to define the correct actions using the Bernoulli Triangle (Fig. \ref{exttriangle}).
\subsubsection{P's correctness} Given $(k,q,n)$, correct actions exist only if $n\le N(k,q)$. In this case, all testing points in the range $[n-N(k,q-1),N(k-1,q-1)]$ are acceptable. Otherwise, there is no correct action.
\subsubsection{OP's correctness} Given $(q,k,n,m)$, When $n>N(k,q)$, any action is regarded as correct if $N(k-1,q-1)\le m\le n-N(k,q-1)$, otherwise, the OP should take ``not break'' if $m>n-N(k,q-1)$ and ``break' if $m<N(k-1,q-1)$; when $n\le N(k,q)$, the OP should take the action ``not break'' if $m<n-N(k,q-1)$ and take action ``break'' if $m>N(k-1,q-1)$. Otherwise, there is no correct action.

\subsection{Complete Game}
\begin{figure}[ht]
\centering
\includegraphics[width=0.4\textwidth]{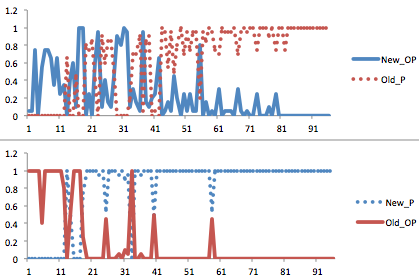}
\caption{Correctness ratio measured for the proposal game on $k=7,q=7$. The legend ``New\_OP'' means that the newly trained neural network plays as an OP; ``Old\_P'' means that the previously trained neural network plays as a P. The same for the following graphs.}
\label{77p}
\end{figure}

\begin{figure}[ht]
\centering
\includegraphics[width=0.4\textwidth]{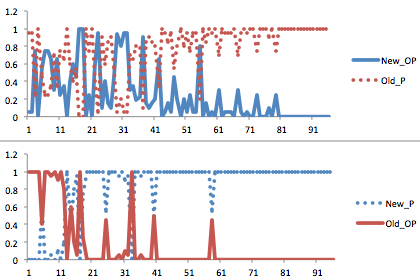}
\caption{Correctness ratio measured for the refutation game on $k=7,q=7$.}
\label{77r}
\end{figure}

In this experiment, we run a full Zermelo game under the given resources $k=7,q=7$. Since there are two neural networks which learn the proposal game and the refutation game respectively, we measure the correctness separately: Fig. \ref{77p} shows the ratio of correctness for each player during the proposal game. And Fig. \ref{77r} shows the ratio of correctness during the refutation game. The horizontal axis is the number of iterations, and it can be seen that the correctness converges extremely slow (80 iterations). It is because, for $k=7,q=7$, the game is relatively complex and the neural MCTS needs more time to find the optimal policy.  
\subsection{Refutation Game}
\begin{figure}[ht]
\centering
\includegraphics[width=0.4\textwidth]{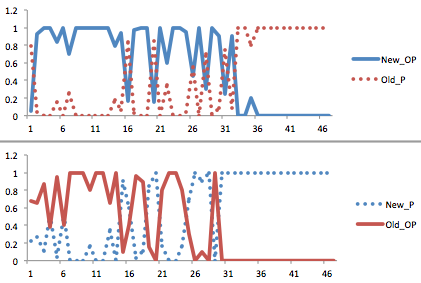}
\caption{Refutation game on $k=7,q=7,n=128$}
\label{77n128}
\end{figure}

\begin{figure}[ht]
\centering
\includegraphics[width=0.4\textwidth]{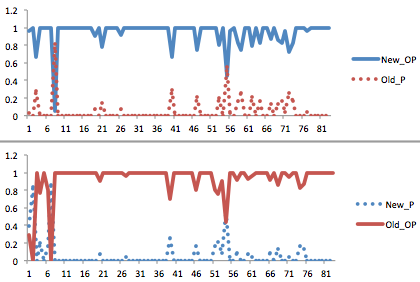}
\caption{Refutation game on $k=7,q=7,n=129$. Notice that in this game, the P is doomed for there is no winning strategy exists.}
\label{77n129}
\end{figure}

To test our method further, we focus our experiment only on refutation games with a given $n$. We first run the experiment on an extreme case where $k=7,q=7$. Using the Bernoulli Triangle (Fig. \ref{exttriangle}), we know that $N(7,7) = 2^7$. We set $n=N(k,q)$ so that the learning process will converge when the P has figured out the optimal winning strategy which is binary search: namely, the first testing point is $2^6$ then $2^5$, $2^4$ and so on. Fig. \ref{77n128} verified that the result is as expected. Then, to study the behavior of our agents under extreme conditions, we run the same experiment on a resource-insufficient case where we keep $k,q$ unchanged and set $n=N(k,q)+1$. In this case, theoretically, no solution exists. Fig. \ref{77n129}, again, verified our expectation and one can see that the P can never find any winning strategy no matter how many iterations it has learned.

\begin{figure}[ht]
\centering
\includegraphics[width=0.4\textwidth]{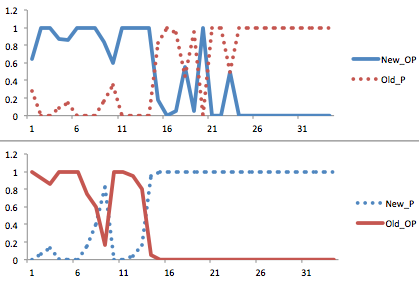}
\caption{Refutation game on $k=3,q=7,n=64$}
\label{37n64}
\end{figure}

\begin{figure}[ht]
\centering
\includegraphics[width=0.4\textwidth]{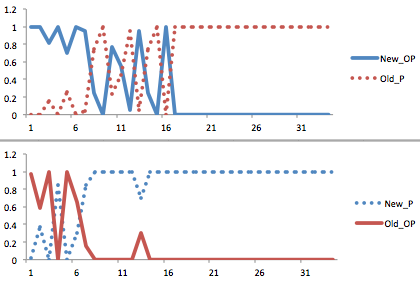}
\caption{Refutation game on $k=3,q=7,n=63$}
\label{37n63}
\end{figure}

In later experiments, we have also tested our method in two more general cases where $k=3,q=7$ for $n=N(3,7)$ (Fig. \ref{37n64}) and $n=N(3,7)-1$ (Fig. \ref{37n63}). All experimental results are conforming to the ground-truth as expected.

The $HSR_{k,q}$ game is also intrinsically asymmetric in terms of training/learning because the OP always takes the last step before the end of the game. This fact makes the game harder to learn for the P. Specifically, considering all possible consequences (in the view of the P) of the last action, there are only three cases: win-win, win-lose, and lose-lose. The OP will lose the game if and only if the consequence is win-win. If the portion of such type of result is tiny, then the OP could exclusively focus on learning the last step while ignoring other steps. However, the P has to learn every step to avoid possible paths which lead him to either win-lose or lose-lose, which, theoretically, are more frequently encountered in the end game. 

\section{Discussion}\label{discuss}
\subsection{State Space Coverage}
\begin{figure}[ht]
\centering
\includegraphics[width=0.4\textwidth]{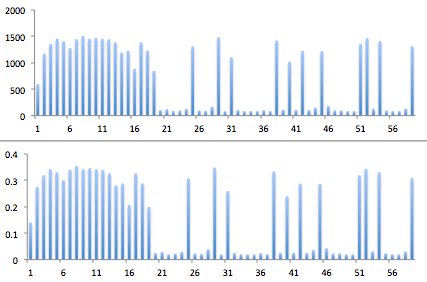}
\caption{States accessed (top) and state space coverage ratio (bottom) during self-play for each iteration, in refutation game $k=7,q=7,n=128$.}
\label{coverage77_128}
\end{figure}

Neural MCTS is capable of handling a large state space \cite{alpha0}. It is necessary for such an algorithm to search only a small portion of the state space and make the decisions on those limited observations. To measure the state space coverage ratio, we recorded the number of states accessed during the experiment, Specifically, in the refutation game $k=7,q=7,n=128$, we count the total number of states accessed during each self-play, and we compute the average state accessed for all 100 self-plays in each iteration. It can be seen in Fig. \ref{coverage77_128} that the maximum number of state accessed is roughly 1500 or 35\% (we have also computed the total number of possible states in this game, which is 4257). As indicated in Fig. \ref{coverage77_128}, at the beginning of the learning, neural MCTS accessed a large number of states, however, once the learning converged, it looked at a few numbers of state and pruned all other irrelevant states. It can also be seen that the coverage ratio will bounce back sometimes, which is due to the exploration during self-play. Our experimental result indicates that changes in coverage ratio might be evidence of adaptive self-pruning in a neural MCTS algorithm, which can be regarded as a justification of its capability of handling large state spaces.

\subsection{Perfectness} This discussion is in the context where the ground truth is known.
Since the correct solution is derived from the optimal policy, it is important to question whether the players are perfect after the training converged (i.e., the correctness of each player becomes flat without further changes). The experimental result shows that, after convergence, for a problem which has a solution, the P always keeps 100\% correctness while the OP rests at 0\%. On the other hand, for a problem which has no solution, the opposite happens. Notice that a consistent 100\% correctness indicates that the player is perfect because, otherwise, the other player will quickly find out the weakness in her adversary. However, there is no guarantee that a consistent 0\% correctness player is also perfect. Since after one player becoming perfect, the other one will always lose no matter what decisions have been made. In this case, all game results are the same, and there is no reward to be gained from further training. Even though, from our experimental observation, the doomed loser is still a robust sub-optimal player after being competitively trained from tabula rasa. The question of when to stop training and how to guarantee that both P and OP become perfect are further topics for future research.

\subsection{Asymmetry}

One can observe some asymmetry in the charts we presented in section \ref{experiment}, and notice that it is always the case that during the beginning iterations the OP is dominating until the P has gained enough experience and learned enough knowledge. Two facts cause this asymmetry: 1. the action space of the P is entirely different from the one of the OP. 2. the OP always takes the last step before the end of the game. These two facts make the game harder to learn for the P but easier for the OP.

\subsection{Limitations}
Our neural MCTS algorithm is time-consuming. It usually takes a large amount of time to converge and we have to use more resources (more CPUs, distributed parallel computing) to make it run faster. That's the reason why we don't experience the amazing performance of AlphaZero for Chess and Go on huge game trees. Another limitation is that to learn the correct action in a discrete action space; the neural MCTS algorithm has to explore all possible actions before learning the correct action. This fact makes the action space a limitation to MCTS like algorithms: the larger the action space, the lower the efficiency of the algorithm.

\section{Future work}
As we have mentioned, $HSR$ is only a prototype for us to apply neural MCTS to problems in other domains. It is still unknown to us whether neural MCTS can be used to solve more complex problems. Our next plan is to try neural MCTS on Quantified Boolean Formulas (QBFs), which is considered to be PSPACE complexity. Since it is quite a natural way to turn a solving process of a QBF into gameplay, we think it could be another touchstone to the capability of neural MCTS. However, since the fundamental symmetry among those QBFs, we plan to turn QBFs into graphs and use a graph neural network to embed them so that symmetry would not be an issue.

\section{Related Work}\label{work}
Imagination-Augmented Agents (I2As \cite{i2as}), an algorithm invented by DeepMind, is used to handle complex games with sparse rewards. Although the algorithm has performed well, it is not model-free. Namely, one has to train, in a supervised way, an imperfect but adequate model first, then use that model to boost the learning process of a regular model-free agent. Even though I2As, along with a trained model, can solve games like Sokoban to some level, I2As can hardly be applied to games where even the training data is limited and hard to generate or label.

By formulating a combinatorial problem as an MDP, Ranked reward \cite{laterre2018ranked} binarized the final reward of an MDP based on a certain threshold, and improves the threshold after each training episode so that the performance is forced to increase during each iteration. However, this method can hardly be applied to problems that already have a binary reward (such as a zero-sum game with reward ${-1,1}$). Even though, the idea that improves the performance threshold after each learning iteration has also been used in AlphaZero as well as our implementation.  

Pointer networks \cite{NIPS2015_5866} have been shown to solve specific combinatorial NP problems with a limited size. The algorithm is based on supervised attention learning on a sequence to sequence RNN. However, due to its high dependency on the quality of data labels (which could be very expensive to obtain), Bello et al. \cite{bello2016neural} improved the method of \cite{NIPS2015_5866} to the RL style. Specifically, they applied actor-critic learning where the actor is the original pointer network, but the critic is a simple REINFORCE \cite{reinforce} style policy gradient. Their result shows a significant improvement in performance. However, this approach can only be applied to sequence decision problem (namely, what is the optimal sequence to finish a task). Also, scalability is still a challenge.

Graph neural networks (GNNs) \cite{battaglia2018relational} are a relatively new approach to hard combinatorial problems. Since some NP-complete problems can be reduced to graph problems, GNNs can capture the internal relational structure efficiently through the message passing process \cite{gilmer2017neural}. Based on message passing and GNNs, Selsam et al. developed a supervised SAT solver: neuroSAT \cite{selsam2018learning}. It has been shown that neuroSAT performs very well on NP-complete problems within a certain size. Combining such GNNs with RL \cite{khalil2017learning} could also be a potential future work direction for us.

\section{Conclusion}\label{conclusion}
Can the amazing game playing capabilities of the neural MCTS algorithm used in AlphaZero for Chess and Go be applied to Zermelo games that have practical significance? We provide a partial positive answer to this question for a class of combinatorial optimization problems which includes the $HSR$ problem. We show how to use \ZG\ (ZG) to translate certain combinatorial optimization problems into Zermelo games: We formulate the optimization problem using predicate logic (where the types of the variables are not "too" complex) and then we use the corresponding semantic game \cite{semgame} as the Zermelo game which we give to the adapted neural MCTS algorithm. For our proof-of-concept example, $HSR$ \ZG, we notice that the 
Zermelo game is asymmetric.

Nevertheless, the adapted neural MCTS algorithm converges on small instances that can be handled by our hardware and finds the winning strategy (and not just an approximation). Our evaluation counts all correct/incorrect moves of the players, thanks to a formal $HSR$ solution we have in the form of the Bernoulli triangle which provides the winning strategy. Besides, we discussed the coverage ratio and transfer learning of our algorithm. We hope our research sheds some light on why the neural MCTS works so well on certain games. While \ZG\ currently is a manual process we hope that many aspects of it can be automated.

{\bf Acknowledgements}: We would like to thank Tal Puhov for his feedback on our paper.

%% file: sample-aamas19.bbl

\begin{thebibliography}{00}


\ifx \showCODEN    \undefined \def \showCODEN     #1{\unskip}     \fi
\ifx \showDOI      \undefined \def \showDOI       #1{#1}\fi
\ifx \showISBNx    \undefined \def \showISBNx     #1{\unskip}     \fi
\ifx \showISBNxiii \undefined \def \showISBNxiii  #1{\unskip}     \fi
\ifx \showISSN     \undefined \def \showISSN      #1{\unskip}     \fi
\ifx \showLCCN     \undefined \def \showLCCN      #1{\unskip}     \fi
\ifx \shownote     \undefined \def \shownote      #1{#1}          \fi
\ifx \showarticletitle \undefined \def \showarticletitle #1{#1}   \fi
\ifx \showURL      \undefined \def \showURL       {\relax}        \fi
\providecommand\bibfield[2]{#2}
\providecommand\bibinfo[2]{#2}
\providecommand\natexlab[1]{#1}
\providecommand\showeprint[2][]{arXiv:#2}

\bibitem[\protect\citeauthoryear{Auer, Cesa-Bianchi, and Fischer}{Auer
  et~al\mbox{.}}{2002}]%
        {ucb1}
\bibfield{author}{\bibinfo{person}{P. Auer}, \bibinfo{person}{N. Cesa-Bianchi},
  {and} \bibinfo{person}{P. Fischer}.} \bibinfo{year}{2002}\natexlab{}.
\newblock \showarticletitle{{Finite-time} {Analysis} of {The} {Multiarmed}
  {Bandit} {Problem}}.
\newblock \bibinfo{journal}{{\em Machine learning\/}} \bibinfo{volume}{47},
  \bibinfo{number}{2} (\bibinfo{year}{2002}), \bibinfo{pages}{235--256}.
\newblock


\bibitem[\protect\citeauthoryear{Battaglia, Hamrick, Bapst, Sanchez-Gonzalez,
  Zambaldi, Malinowski, Tacchetti, Raposo, Santoro, Faulkner,
  et~al\mbox{.}}{Battaglia et~al\mbox{.}}{2018}]%
        {battaglia2018relational}
\bibfield{author}{\bibinfo{person}{Peter~W Battaglia},
  \bibinfo{person}{Jessica~B Hamrick}, \bibinfo{person}{Victor Bapst},
  \bibinfo{person}{Alvaro Sanchez-Gonzalez}, \bibinfo{person}{Vinicius
  Zambaldi}, \bibinfo{person}{Mateusz Malinowski}, \bibinfo{person}{Andrea
  Tacchetti}, \bibinfo{person}{David Raposo}, \bibinfo{person}{Adam Santoro},
  \bibinfo{person}{Ryan Faulkner}, {et~al\mbox{.}}}
  \bibinfo{year}{2018}\natexlab{}.
\newblock \showarticletitle{Relational inductive biases, deep learning, and
  graph networks}.
\newblock \bibinfo{journal}{{\em arXiv preprint arXiv:1806.01261\/}}
  (\bibinfo{year}{2018}).
\newblock


\bibitem[\protect\citeauthoryear{Bello, Pham, Le, Norouzi, and Bengio}{Bello
  et~al\mbox{.}}{2016}]%
        {bello2016neural}
\bibfield{author}{\bibinfo{person}{Irwan Bello}, \bibinfo{person}{Hieu Pham},
  \bibinfo{person}{Quoc~V Le}, \bibinfo{person}{Mohammad Norouzi}, {and}
  \bibinfo{person}{Samy Bengio}.} \bibinfo{year}{2016}\natexlab{}.
\newblock \showarticletitle{Neural combinatorial optimization with
  reinforcement learning}.
\newblock \bibinfo{journal}{{\em arXiv preprint arXiv:1611.09940\/}}
  (\bibinfo{year}{2016}).
\newblock


\bibitem[\protect\citeauthoryear{Browne, Powley, Whitehouse, Lucas, Cowling,
  Rohlfshagen, Tavener, Liebana, Samothrakis, and Colton}{Browne
  et~al\mbox{.}}{2012}]%
        {mcts_survey}
\bibfield{author}{\bibinfo{person}{Cameron Browne},
  \bibinfo{person}{Edward~Jack Powley}, \bibinfo{person}{Daniel Whitehouse},
  \bibinfo{person}{Simon~M. Lucas}, \bibinfo{person}{Peter~I. Cowling},
  \bibinfo{person}{Philipp Rohlfshagen}, \bibinfo{person}{Stephen Tavener},
  \bibinfo{person}{Diego~Perez Liebana}, \bibinfo{person}{Spyridon
  Samothrakis}, {and} \bibinfo{person}{Simon Colton}.}
  \bibinfo{year}{2012}\natexlab{}.
\newblock \showarticletitle{{A} {Survey} of {Monte} {Carlo} {Tree} {Search}
  {Methods}.}
\newblock \bibinfo{journal}{{\em IEEE Trans. Comput. Intellig. and AI in
  Games\/}} \bibinfo{volume}{4}, \bibinfo{number}{1} (\bibinfo{year}{2012}),
  \bibinfo{pages}{1--43}.
\newblock


\bibitem[\protect\citeauthoryear{Gilmer, Schoenholz, Riley, Vinyals, and
  Dahl}{Gilmer et~al\mbox{.}}{2017}]%
        {gilmer2017neural}
\bibfield{author}{\bibinfo{person}{Justin Gilmer}, \bibinfo{person}{Samuel~S
  Schoenholz}, \bibinfo{person}{Patrick~F Riley}, \bibinfo{person}{Oriol
  Vinyals}, {and} \bibinfo{person}{George~E Dahl}.}
  \bibinfo{year}{2017}\natexlab{}.
\newblock \showarticletitle{Neural message passing for quantum chemistry}. In
  \bibinfo{booktitle}{{\em Proceedings of the 34th International Conference on
  Machine Learning-Volume 70}}. JMLR. org, \bibinfo{pages}{1263--1272}.
\newblock


\bibitem[\protect\citeauthoryear{Hintikka}{Hintikka}{1982}]%
        {semgame}
\bibfield{author}{\bibinfo{person}{Jaakko Hintikka}.}
  \bibinfo{year}{1982}\natexlab{}.
\newblock \showarticletitle{Game-theoretical semantics: insights and
  prospects.}
\newblock \bibinfo{journal}{{\em Notre Dame J. Formal Logic\/}}
  \bibinfo{volume}{23}, \bibinfo{number}{2} (\bibinfo{date}{04}
  \bibinfo{year}{1982}), \bibinfo{pages}{219--241}.
\newblock


\bibitem[\protect\citeauthoryear{Khalil, Dai, Zhang, Dilkina, and Song}{Khalil
  et~al\mbox{.}}{2017}]%
        {khalil2017learning}
\bibfield{author}{\bibinfo{person}{Elias Khalil}, \bibinfo{person}{Hanjun Dai},
  \bibinfo{person}{Yuyu Zhang}, \bibinfo{person}{Bistra Dilkina}, {and}
  \bibinfo{person}{Le Song}.} \bibinfo{year}{2017}\natexlab{}.
\newblock \showarticletitle{Learning combinatorial optimization algorithms over
  graphs}. In \bibinfo{booktitle}{{\em Advances in Neural Information
  Processing Systems}}. \bibinfo{pages}{6348--6358}.
\newblock


\bibitem[\protect\citeauthoryear{Kocsis and Szepesvari}{Kocsis and
  Szepesvari}{2006}]%
        {uct}
\bibfield{author}{\bibinfo{person}{Levente Kocsis} {and} \bibinfo{person}{Csaba
  Szepesvari}.} \bibinfo{year}{2006}\natexlab{}.
\newblock \showarticletitle{{Bandit} {Based} {Monte-Carlo} {Planning}.}. In
  \bibinfo{booktitle}{{\em ECML}} {\em (\bibinfo{series}{Lecture Notes in
  Computer Science})}, Vol.~\bibinfo{volume}{4212}.
  \bibinfo{publisher}{Springer}, \bibinfo{pages}{282--293}.
\newblock


\bibitem[\protect\citeauthoryear{Laterre, Fu, Jabri, Cohen, Kas, Hajjar, Dahl,
  Kerkeni, and Beguir}{Laterre et~al\mbox{.}}{2018}]%
        {laterre2018ranked}
\bibfield{author}{\bibinfo{person}{Alexandre Laterre}, \bibinfo{person}{Yunguan
  Fu}, \bibinfo{person}{Mohamed~Khalil Jabri}, \bibinfo{person}{Alain-Sam
  Cohen}, \bibinfo{person}{David Kas}, \bibinfo{person}{Karl Hajjar},
  \bibinfo{person}{Torbjorn~S Dahl}, \bibinfo{person}{Amine Kerkeni}, {and}
  \bibinfo{person}{Karim Beguir}.} \bibinfo{year}{2018}\natexlab{}.
\newblock \showarticletitle{Ranked Reward: Enabling Self-Play Reinforcement
  Learning for Combinatorial Optimization}.
\newblock \bibinfo{journal}{{\em arXiv preprint arXiv:1807.01672\/}}
  (\bibinfo{year}{2018}).
\newblock


\bibitem[\protect\citeauthoryear{Mnih, Kavukcuoglu, Silver, Rusu, Veness,
  Bellemare, Graves, Riedmiller, Fidjeland, Ostrovski, Petersen, Beattie,
  Sadik, Antonoglou, King, Kumaran, Wierstra, Legg, and Hassabis}{Mnih
  et~al\mbox{.}}{2015}]%
        {dqn}
\bibfield{author}{\bibinfo{person}{Volodymyr Mnih}, \bibinfo{person}{Koray
  Kavukcuoglu}, \bibinfo{person}{David Silver}, \bibinfo{person}{Andrei~A.
  Rusu}, \bibinfo{person}{Joel Veness}, \bibinfo{person}{Marc~G. Bellemare},
  \bibinfo{person}{Alex Graves}, \bibinfo{person}{Martin Riedmiller},
  \bibinfo{person}{Andreas~K. Fidjeland}, \bibinfo{person}{Georg Ostrovski},
  \bibinfo{person}{Stig Petersen}, \bibinfo{person}{Charles Beattie},
  \bibinfo{person}{Amir Sadik}, \bibinfo{person}{Ioannis Antonoglou},
  \bibinfo{person}{Helen King}, \bibinfo{person}{Dharshan Kumaran},
  \bibinfo{person}{Daan Wierstra}, \bibinfo{person}{Shane Legg}, {and}
  \bibinfo{person}{Demis Hassabis}.} \bibinfo{year}{2015}\natexlab{}.
\newblock \showarticletitle{Human-level control through deep reinforcement
  learning}.
\newblock \bibinfo{journal}{{\em Nature\/}} \bibinfo{volume}{518},
  \bibinfo{number}{7540} (\bibinfo{date}{Feb.} \bibinfo{year}{2015}),
  \bibinfo{pages}{529--533}.
\newblock
\showISSN{00280836}


\bibitem[\protect\citeauthoryear{Rosin}{Rosin}{2011}]%
        {Rosin2011}
\bibfield{author}{\bibinfo{person}{Christopher~D. Rosin}.}
  \bibinfo{year}{2011}\natexlab{}.
\newblock \showarticletitle{Multi-armed bandits with episode context}.
\newblock \bibinfo{journal}{{\em Annals of Mathematics and Artificial
  Intelligence\/}} \bibinfo{volume}{61}, \bibinfo{number}{3}
  (\bibinfo{date}{mar} \bibinfo{year}{2011}), \bibinfo{pages}{203--230}.
\newblock


\bibitem[\protect\citeauthoryear{Schulman, Wolski, Dhariwal, Radford, and
  Klimov}{Schulman et~al\mbox{.}}{2017}]%
        {ppo}
\bibfield{author}{\bibinfo{person}{John Schulman}, \bibinfo{person}{Filip
  Wolski}, \bibinfo{person}{Prafulla Dhariwal}, \bibinfo{person}{Alec Radford},
  {and} \bibinfo{person}{Oleg Klimov}.} \bibinfo{year}{2017}\natexlab{}.
\newblock \bibinfo{title}{Proximal Policy Optimization Algorithms}.
\newblock   (\bibinfo{year}{2017}).
\newblock
\showeprint{arXiv:1707.06347}


\bibitem[\protect\citeauthoryear{Selsam, Lamm, Bunz, Liang, de~Moura, and
  Dill}{Selsam et~al\mbox{.}}{2018}]%
        {selsam2018learning}
\bibfield{author}{\bibinfo{person}{Daniel Selsam}, \bibinfo{person}{Matthew
  Lamm}, \bibinfo{person}{Benedikt Bunz}, \bibinfo{person}{Percy Liang},
  \bibinfo{person}{Leonardo de Moura}, {and} \bibinfo{person}{David~L Dill}.}
  \bibinfo{year}{2018}\natexlab{}.
\newblock \showarticletitle{Learning a SAT Solver from Single-Bit Supervision}.
\newblock \bibinfo{journal}{{\em arXiv preprint arXiv:1802.03685\/}}
  (\bibinfo{year}{2018}).
\newblock


\bibitem[\protect\citeauthoryear{Silver, Huang, Maddison, Guez, Sifre, van~den
  Driessche, Schrittwieser, Antonoglou, Panneershelvam, Lanctot, Dieleman,
  Grewe, Nham, Kalchbrenner, Sutskever, Lillicrap, Leach, Kavukcuoglu, Graepel,
  and Hassabis}{Silver et~al\mbox{.}}{2016}]%
        {alpha_go}
\bibfield{author}{\bibinfo{person}{David Silver}, \bibinfo{person}{Aja Huang},
  \bibinfo{person}{Chris~J. Maddison}, \bibinfo{person}{Arthur Guez},
  \bibinfo{person}{Laurent Sifre}, \bibinfo{person}{George van~den Driessche},
  \bibinfo{person}{Julian Schrittwieser}, \bibinfo{person}{Ioannis Antonoglou},
  \bibinfo{person}{Veda Panneershelvam}, \bibinfo{person}{Marc Lanctot},
  \bibinfo{person}{Sander Dieleman}, \bibinfo{person}{Dominik Grewe},
  \bibinfo{person}{John Nham}, \bibinfo{person}{Nal Kalchbrenner},
  \bibinfo{person}{Ilya Sutskever}, \bibinfo{person}{Timothy Lillicrap},
  \bibinfo{person}{Madeleine Leach}, \bibinfo{person}{Koray Kavukcuoglu},
  \bibinfo{person}{Thore Graepel}, {and} \bibinfo{person}{Demis Hassabis}.}
  \bibinfo{year}{2016}\natexlab{}.
\newblock \showarticletitle{{Mastering} the game of {Go} with deep neural
  networks and tree search}.
\newblock \bibinfo{journal}{{\em Nature\/}}  \bibinfo{volume}{529}
  (\bibinfo{date}{Jan.} \bibinfo{year}{2016}), \bibinfo{pages}{484}.
\newblock


\bibitem[\protect\citeauthoryear{Silver, Hubert, Schrittwieser, Antonoglou,
  Lai, Guez, Lanctot, Sifre, Kumaran, Graepel, et~al\mbox{.}}{Silver
  et~al\mbox{.}}{2018}]%
        {silver2018general}
\bibfield{author}{\bibinfo{person}{David Silver}, \bibinfo{person}{Thomas
  Hubert}, \bibinfo{person}{Julian Schrittwieser}, \bibinfo{person}{Ioannis
  Antonoglou}, \bibinfo{person}{Matthew Lai}, \bibinfo{person}{Arthur Guez},
  \bibinfo{person}{Marc Lanctot}, \bibinfo{person}{Laurent Sifre},
  \bibinfo{person}{Dharshan Kumaran}, \bibinfo{person}{Thore Graepel},
  {et~al\mbox{.}}} \bibinfo{year}{2018}\natexlab{}.
\newblock \showarticletitle{A general reinforcement learning algorithm that
  masters Chess, Shogi, and Go through self-play}.
\newblock \bibinfo{journal}{{\em Science\/}} \bibinfo{volume}{362},
  \bibinfo{number}{6419} (\bibinfo{year}{2018}), \bibinfo{pages}{1140--1144}.
\newblock


\bibitem[\protect\citeauthoryear{Silver, Hubert, Schrittwieser, Antonoglou,
  Lai, Guez, Lanctot, Sifre, Kumaran, Graepel, Lillicrap, Simonyan, and
  Hassabis}{Silver et~al\mbox{.}}{2017a}]%
        {gochessshogi}
\bibfield{author}{\bibinfo{person}{David Silver}, \bibinfo{person}{Thomas
  Hubert}, \bibinfo{person}{Julian Schrittwieser}, \bibinfo{person}{Ioannis
  Antonoglou}, \bibinfo{person}{Matthew Lai}, \bibinfo{person}{Arthur Guez},
  \bibinfo{person}{Marc Lanctot}, \bibinfo{person}{Laurent Sifre},
  \bibinfo{person}{Dharshan Kumaran}, \bibinfo{person}{Thore Graepel},
  \bibinfo{person}{Timothy~P. Lillicrap}, \bibinfo{person}{Karen Simonyan},
  {and} \bibinfo{person}{Demis Hassabis}.} \bibinfo{year}{2017}\natexlab{a}.
\newblock \showarticletitle{{Mastering} {Chess} and {Shogi} by {Self-Play} with
  a {General} {Reinforcement} {Learning} {Algorithm}.}
\newblock \bibinfo{journal}{{\em CoRR\/}}  \bibinfo{volume}{abs/1712.01815}
  (\bibinfo{year}{2017}).
\newblock


\bibitem[\protect\citeauthoryear{Silver, Schrittwieser, Simonyan, Antonoglou,
  Huang, Guez, Hubert, Baker, Lai, Bolton, Chen, Lillicrap, Hui, Sifre, van~den
  Driessche, Graepel, and Hassabis}{Silver et~al\mbox{.}}{2017b}]%
        {alpha0}
\bibfield{author}{\bibinfo{person}{David Silver}, \bibinfo{person}{Julian
  Schrittwieser}, \bibinfo{person}{Karen Simonyan}, \bibinfo{person}{Ioannis
  Antonoglou}, \bibinfo{person}{Aja Huang}, \bibinfo{person}{Arthur Guez},
  \bibinfo{person}{Thomas Hubert}, \bibinfo{person}{Lucas Baker},
  \bibinfo{person}{Matthew Lai}, \bibinfo{person}{Adrian Bolton},
  \bibinfo{person}{Yutian Chen}, \bibinfo{person}{Timothy Lillicrap},
  \bibinfo{person}{Fan Hui}, \bibinfo{person}{Laurent Sifre},
  \bibinfo{person}{George van~den Driessche}, \bibinfo{person}{Thore Graepel},
  {and} \bibinfo{person}{Demis Hassabis}.} \bibinfo{year}{2017}\natexlab{b}.
\newblock \showarticletitle{{Mastering} the game of {Go} without human
  knowledge}.
\newblock \bibinfo{journal}{{\em Nature\/}}  \bibinfo{volume}{550}
  (\bibinfo{date}{Oct.} \bibinfo{year}{2017}), \bibinfo{pages}{354}.
\newblock


\bibitem[\protect\citeauthoryear{Vinyals, Fortunato, and Jaitly}{Vinyals
  et~al\mbox{.}}{2015}]%
        {NIPS2015_5866}
\bibfield{author}{\bibinfo{person}{Oriol Vinyals}, \bibinfo{person}{Meire
  Fortunato}, {and} \bibinfo{person}{Navdeep Jaitly}.}
  \bibinfo{year}{2015}\natexlab{}.
\newblock \showarticletitle{Pointer Networks}.
\newblock In \bibinfo{booktitle}{{\em Advances in Neural Information Processing
  Systems 28}}, \bibfield{editor}{\bibinfo{person}{C.~Cortes},
  \bibinfo{person}{N.~D. Lawrence}, \bibinfo{person}{D.~D. Lee},
  \bibinfo{person}{M.~Sugiyama}, {and} \bibinfo{person}{R.~Garnett}} (Eds.).
  \bibinfo{publisher}{Curran Associates, Inc.}, \bibinfo{pages}{2692--2700}.
\newblock


\bibitem[\protect\citeauthoryear{Weber, Racaniere, Reichert, Buesing, Guez,
  Rezende, Badia, Vinyals, Heess, Li, Pascanu, Battaglia, Hassabis, Silver, and
  Wierstra}{Weber et~al\mbox{.}}{2017}]%
        {i2as}
\bibfield{author}{\bibinfo{person}{Theophane Weber}, \bibinfo{person}{Sebastien
  Racaniere}, \bibinfo{person}{David~P. Reichert}, \bibinfo{person}{Lars
  Buesing}, \bibinfo{person}{Arthur Guez}, \bibinfo{person}{Danilo~Jimenez
  Rezende}, \bibinfo{person}{Adria~Puigdomenech Badia}, \bibinfo{person}{Oriol
  Vinyals}, \bibinfo{person}{Nicolas Heess}, \bibinfo{person}{Yujia Li},
  \bibinfo{person}{Razvan Pascanu}, \bibinfo{person}{Peter Battaglia},
  \bibinfo{person}{Demis Hassabis}, \bibinfo{person}{David Silver}, {and}
  \bibinfo{person}{Daan Wierstra}.} \bibinfo{year}{2017}\natexlab{}.
\newblock \bibinfo{title}{Imagination-Augmented Agents for Deep Reinforcement
  Learning}.
\newblock   (\bibinfo{year}{2017}).
\newblock
\showeprint{arXiv:1707.06203}


\bibitem[\protect\citeauthoryear{Williams}{Williams}{1992}]%
        {reinforce}
\bibfield{author}{\bibinfo{person}{R.~J. Williams}.}
  \bibinfo{year}{1992}\natexlab{}.
\newblock \showarticletitle{Simple statistical gradient-following algorithms
  for connectionist reinforcement learning}.
\newblock \bibinfo{journal}{{\em Machine Learning\/}}  \bibinfo{volume}{8}
  (\bibinfo{year}{1992}), \bibinfo{pages}{229--256}.
\newblock


\end{thebibliography}
